\pdfoutput=1

\documentclass[11pt]{article}

\usepackage{emnlp2022}

\usepackage{times}
\usepackage{latexsym}

\usepackage[T1]{fontenc}

\usepackage[utf8]{inputenc}

\usepackage{microtype}

\usepackage{booktabs}
\usepackage{tabularx}
\usepackage{xcolor}

\newcommand{\dataset}{SQuALITY}

\usepackage{multirow}
\usepackage{graphicx}
\usepackage{makecell}

\usepackage{amsfonts}
\usepackage{pifont}
\newcommand*\colorcmark[1]{%
  \expandafter\newcommand\csname #1check\endcsname{\textcolor{#1}{\ding{51}}}%
}
\newcommand*\colorxmark[1]{%
  \expandafter\newcommand\csname #1xmark\endcsname{\textcolor{#1}{\ding{55}}}%
}
\colorcmark{green}
\colorxmark{red}

\title{SQuALITY: Building a Long-Document Summarization Dataset \\ the Hard Way}

\author{Alex Wang~~~Richard Yuanzhe Pang~~~Angelica Chen~~~Jason Phang~~~Samuel R. Bowman \\
New York University\\
{\tt alexwang@nyu.edu~~~yzpang@nyu.edu~~~bowman@nyu.edu}}

\begin{document}
\maketitle
\begin{abstract}
Summarization datasets are often assembled either by scraping naturally occurring public-domain summaries---which are nearly always in difficult-to-work-with technical domains---or by using approximate heuristics to extract them from everyday text---which frequently yields unfaithful summaries.
In this work, we turn to a slower but more straightforward approach to developing summarization benchmark data: We hire highly-qualified contractors to read stories and write original summaries from scratch. To amortize reading time, we collect five summaries per document, with the first giving an overview and the subsequent four addressing specific questions.
We use this protocol to collect SQuALITY, a dataset of question-focused summaries built on the same public-domain short stories as the multiple-choice dataset QuALITY \citep{pang2021quality}.
Experiments with state-of-the-art summarization systems show that our dataset is challenging 
and that existing automatic evaluation metrics are weak indicators of quality.
\end{abstract}

\begin{figure}[ht!]
     \centering
     \includegraphics[width=0.99\columnwidth]{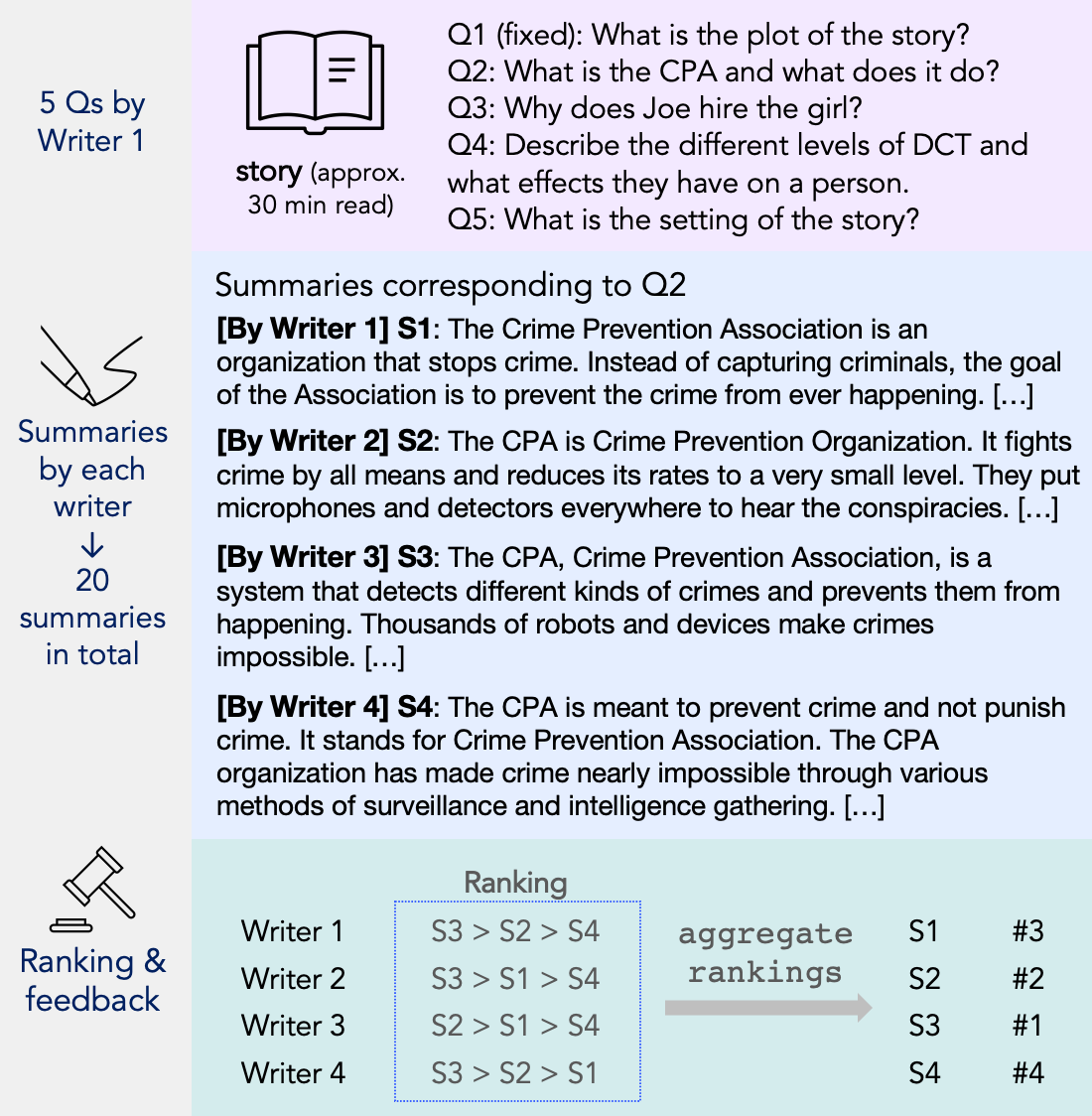}
    \caption{An overview of our data collection pipeline. One writer first creates four questions, with an additional fixed question used for every story. Then, four writers each create summaries answering the five questions. Next, each writer ranks the other three summaries for each question and provides written feedback. Finally, we aggregate ranks and award bonuses to incentivize high-quality summaries and careful feedback. Between data collection rounds, writers review the feedback their summaries received.}
    \label{fig:flowchart}
    \vspace{-0.3cm}
\end{figure}

\section{Introduction}

Research on automatic text summarization depends upon the availability of adequate benchmark datasets. Existing datasets in this area often have issues that seriously limit their usability:
For instance, summaries from the popular scraped benchmark summarization dataset CNN/DailyMail \citep{nallapati2016abstractive} contain HTML artifacts, links to other news articles, and other types of noise \citep{kryscinski-etal-2019-neural,tejaswin-etal-2021-well}.

A common approach to creating summarization datasets is to develop heuristics to extract pseudo-summaries from existing texts.
While scraped summaries can be cleaned of noise, these heuristics can lead to more fundamental data artifacts.
For example, the XSum dataset \citep{narayan-etal-2018-dont} was created by extracting the first sentence of a news article to act as the summary for the rest of the document.
However, studies have found that 30--50\% of summaries created this way contain facts that are unsupported by the rest of the article \citep{tejaswin-etal-2021-well,nan-etal-2021-entity}.
Models trained on this dataset learn to repeat this noise pattern by hallucinating facts in their outputs.
It appears that known heuristics do not produce reliable data.

Another approach to creating summarization datasets relies on serendipity in finding naturally occurring summaries.
For example, the arXiv and PubMed summarization datasets \citep{cohan-etal-2018-discourse} use the abstracts of scientific papers as summaries of the papers. 
BigPatent \citep{sharma-etal-2019-bigpatent} and GovReport \citep{huang-etal-2021-efficient} use expert-written summaries that come with patent filings and government reports, respectively.
While these summaries are likely high-quality, the domain of the data poses a significant challenge for system evaluation: 
Automatic evaluation metrics for summarization are unreliable \citep{kryscinski-etal-2019-neural,gehrmann2022repairing}, but the summaries are too technical and jargonistic for non-specialist human raters to evaluate reliably either. 
Because we rely on chance in finding these summaries, we are beholden to whatever domain they come from, rather than the domain we are interested in.

Relying on finding and scraping summarization data is also problematic in that, often, the found data is proprietary and not freely distributable.
For example, many researchers and organizations are unwilling to host or distribute the CNN/DailyMail dataset,\footnote{\href{https://github.com/abisee/cnn-dailymail/issues/9}{See discussion here.}} despite it being one of the most popular summarization datasets to experiment on.
Similarly, several recent summarization datasets built on data such as scientific journal papers \citep{meng-etal-2021-bringing} or SparkNotes book summaries \citep{ladhak-etal-2020-exploring,kryscinski2021booksum} have never been made available to researchers, with the dataset creators instead asking potential data users to re-scrape them individually, which can be a serious obstacle to reproducibility.

In this work, we propose a crowdsourcing protocol for collecting original summaries that are free of these issues.
Crowdsourcing summaries has been under-explored because straightforward approaches for doing so is labor-intensive and extremely expensive.
While our protocol is still fairly expensive, we structure it in a way that makes the cost per summary more tractable ($\sim$\$6/summary) while also including incentives and checks to ensure the summaries are high-quality.
The protocol does not rely on finding naturally occurring summaries and is agnostic to the input documents used, so we are free to choose what input documents we want to summarize. 
We use short stories from Project Gutenberg to avoid the aforementioned domain and licensing issues.

We use this protocol to collect SQuALITY\footnote{SQuALITY is related to the multiple choice QA dataset QuALITY \citep{pang2021quality} in that it uses some of the same stories as QuALITY in the same data splits.} (Summarization-format QUestion Answering with Long Input Texts, Yes!), a dataset for question-focused abstractive summarization of short stories.
SQuALITY summaries are created by having trained writers read short stories from Project Gutenberg, then ask questions about different aspects of the story. 
The writers then answer the questions by writing summaries focusing on that aspect.
Each question is answered by four different annotators, who then review each other's work to ensure the data is high-quality.
In total, SQuALITY consists of 100 stories, 500 questions, and 2000 summaries.\footnote{This paper releases SQuALITY v1.0. We will soon release SQuALITY v1.1, which consists of 127 stories.}

Overall, we make the following contributions:
\begin{enumerate}
    \item We develop a crowdsourcing protocol for collecting summaries that partially ameliorates the high cost of crowdsourcing long textual responses while maintaining data quality.
    \item We use this protocol to collect SQuALITY, an abstractive summarization dataset. SQuALITY is question-focused, multi-reference, and distributed with a CC BY license.
    \item We conduct preliminary experiments on SQuALITY with pretrained language models using human evaluation. We find that state-of-the-art summarization models produce summaries that are significantly worse than human-written summaries.
    \item We identify that common automatic evaluation metrics for summarization correlate very poorly with human judgments of quality. We also find that having multiple references when computing automatic evaluation metrics does not improve the correlation of the metric.
\end{enumerate}

SQuALITY is a challenging benchmark for long-context text generation models.
We make SQuALITY, our models, 
and our templates for human evaluation of model outputs available at \url{https://github.com/nyu-mll/SQuALITY}.

\begin{table*}[t]
\small
    \centering
    \begin{tabularx}{\textwidth}{XX}
    \toprule
    \multicolumn{2}{l}{Title: \href{https://www.gutenberg.org/ebooks/51656}{Pick A Crime} (\url{https://www.gutenberg.org/ebooks/51656})} \\ 
    \midrule
    \multicolumn{2}{l}{Q: What is the CPA and what does it do?} \\ 
    \midrule
    The Crime Prevention Association is an organization that stops crime. Instead of capturing criminals, the goal of the Association is to prevent the crime from ever happening. They implement thousands of crime-prevention methods and devices. There are many amateur cops who constantly follow criminals around in hopes of catching them in the act so that they may be hailed a hero and... 
    & 
    The CPA is Crime Prevention Organization. It fights crime by all means and reduces its rates to a very small level. They put microphones and detectors everywhere to hear the conspiracies. They place robots as bartenders to control the level of alcohol in visitors to prevent them being drunk. They make all the women learn self-defense. The organization's made crime almost impossible...
    \\
    \midrule
    The CPA, Crime Prevention Association, is a system that detects different kinds of crimes and prevents them from happening. Thousands of robots and devices make crimes impossible. The association will not punish any crime, instead, the criminal will be send to a CPA hospital for some treatments that will result in getting the best jobs. The CPA also hands out ID cards that states one’s... 
    & 
    The CPA is meant to prevent crime and not punish crime. It stands for Crime Prevention Association. The CPA organization has made crime nearly impossible through various methods of surveillance and intelligence gathering. The crime was not punished by the CPA but addressed by sending the person to a hospital for expensive treatment to correct and remove the deviance from the person’s... 
    \\
    \bottomrule
    \end{tabularx}
    \caption{An example question and four human-written references from SQuALITY. The full references are available in Table~\ref{tab:dataset_examples_long} in the appendix.}
    \label{tab:dataset_examples_short}
\end{table*}

\section{Related Work}
\label{sec:related_work}

\subsection{Story Summarization}

A long-standing line of summarization datasets focuses on summarizing books and stories.
More recently, \citet{kryscinski2021booksum} introduce BookSum, which consists of public domain books and summaries of those books, book chapters, and book paragraphs.
Similarly, \citet{ladhak-etal-2020-exploring} propose a dataset for summarizing extracted text from chapters of public domain books.
The summaries of both these datasets are scraped from popular study guide websites such as SparkNotes, apparently without an overt license, and thus the datasets cannot be legally distributed.
Adjacent to book summarization, \citet{chen2022summscreen} introduce SummScreen, which consists of fan-written transcripts of TV show episodes paired with Wikipedia and fan-written summaries of those episodes.

\subsection{Question-Focused Summarization}

Question-focused summarization (QFS) is a variant of summarization where the summary answers a question about the input document or focuses on a particular aspect of the document.
QFS is especially useful when summarizing long documents, as there are typically many topics and entities mentioned in these documents, and different users can be interested in different topics.
In response to growing interest, several QFS datasets have been recently proposed.

The Debatepedia dataset \citep{nema-etal-2017-diversity} is a found dataset of articles explaining social and philosophical issues.
Each article consists of a background paragraph about the issue, along with a set of questions about the issue and short answers to those questions. 
FacetSum \citep{meng-etal-2021-bringing} is a found dataset consisting of a corpus of scientific papers paired with author-written summaries focusing on different aspects of the paper.
WikiAsp \citep{hayashi-etal-2021-wikiasp} and AQuaMuSe \citep{kulkarni2020aquamuse} are two heuristically created, multi-document QFS datasets derived from Wikipedia.

Most similar to our dataset is QMSum \citep{zhong-etal-2021-qmsum}, a long-document QFS dataset where the input documents are meeting transcripts.
Similar to our work, they crowdsource questions and summaries by having undergraduate students read the full transcripts and write questions about them, guided by a list of prewritten question templates.
Unlike our work, their primary mechanism for quality control is manually reviewing the collected responses, whereas we design a crowdsourcing protocol wherein writers review each other's work.
As a result of our protocol, we collect multiple reference summaries for each question, while they have only one reference summary per question.
Additionally, they distinguish between general and specific questions, where the latter questions can be sometimes be answered with local context, e.g. ``What did $A$ recommend to do when discussing $X$ and why?'', whereas we emphasize questions that draw on the entire story.


\subsection{Long-Form Question Answering}

Question-focused summarization can be considered a special case of long-form question answering (LFQA).
In LFQA, the inputs are similarly a question and an input document, and the task is to produce a ``long'' answer, as opposed to an answer that is a short phrase or span of text.
Qualitatively, we distinguish QFS and LFQA as the summaries for QFS should cover multiple parts of the input document, if not the whole document, whereas LFQA answers can draw from a single portion of the document. 
Additionally, \citet{krishna-etal-2021-hurdles} found that pretrained language models can answer questions from LFQA datasets without utilizing the input documents due to having memorized relevant information during pretraining.
This reliance on memorized information is problematic because the memorized information is not exposed to users and may be out of date or irrelevant to the input document.
For QFS, on the other hand, responses to each question depend heavily on the input document, and in the case of SQuALITY, have never been on the web because the summaries are crowdsourced.
However, the two tasks share many of the same properties and challenges.

\section{Dataset Construction}

Our high-level approach to collecting summaries is to hire writers to create multiple summaries of a long input document with an incentive structure that encourages high-quality summaries by having other hired workers review the quality of the summaries.

\subsection{Source Documents}

Our considerations in selecting a corpus of documents for which to collect summaries are:
(1) The documents are long, as document-level tasks are more challenging than paragraph-level ones;
(2) The documents can support several substantive summaries, as we will collect multiple summaries per document for cost-efficiency (see Section~\ref{sec:crowdsourcing_writing});
(3) The documents have a permissive license such that they can be easily distributed;
(4) The documents are lay-accessible, such that the average college-educated English-fluent speaker can both understand them and confidently evaluate the correctness of summaries derived from them.

We use short stories from Project Gutenberg as they meet all of these desiderata.\footnote{\url{https://www.gutenberg.org/}}
Specifically, we use a collection of science fiction short stories written in the 1930s--1970s.
We select stories that are between 3000 to 6000 words long.
Many of the stories used are also included in the QuALITY \citep{pang2021quality} dataset, and we coordinate with the QuALITY creators such that stories that appear in both datasets are assigned to the same split.
Additionally, we follow the same preprocessing for the stories used in QuALITY.

\subsection{Question-Focused Summarization}
\label{sec:crowdsourcing_writing}

For crowdworkers to write accurate and high-quality summaries, they need to read the entire story, which takes 20--40 minutes.
Rather than asking writers to create one summary per story, our strategy is to collect multiple summaries per story to amortize the cost of reading across summaries.

We consider a variant of the summarization task known as question-focused summarization where summaries are intended to answer a question about the document.
Each story, then, is associated with multiple questions and a writer creates multiple summaries for each story to answer those questions.
Question-focused summarization has received increasing attention from the summarization literature in recent years as a task in its own right (see Section~\ref{sec:related_work} for an overview), and we expect it to be a viable proxy benchmark task for narrative-text summarization more broadly.

\subsection{Writing}

In the first step of the crowdsourcing pipeline, we ask writers to read the story and then create questions satisfying two general criteria.
First, we ask that writers create questions that require the whole or multiple parts of the story to answer, as opposed to a single sentence, paragraph, or span. 
Second, to minimize disagreements in evaluation, we ask writers to avoid questions that speculate substantially beyond the literal text of the story when interpreting themes or symbolism. 
To assist writers in creating questions that satisfy these properties, we provide them with a list of question templates that we expect will satisfy these properties in most cases, shown in Appendix~\ref{ax:question-templates}.
Writers can also write story-specific questions not based on any of these templates so long as they follow the criteria.

For each story, we assign one worker to create four questions. After the questions have been created, they are then answered by four writers, including the original question writer.
We also ask that each writer creates a general story summary, framed as answering the question ``What is the plot of the story?'', for a total of five questions per story.
Responses are required to be between 75 to 500 words long, to avoid copying the text of the story verbatim, and to draw on different parts of the story as much as possible.
Writers report that this process takes 40--120 minutes, including time to read the story.

\subsection{Data Validation} 

After a writing phase, for each story, we have five questions with four reference summaries per question.
In the second phase of the crowdsourcing pipeline, we ask workers to review the responses to ensure they are high-quality.

As with writing, asking crowdworkers to review the collected responses is expensive because verifying whether a response is high-quality and faithful to the story requires the reviewer to have read the entire story.
Our strategy to minimize costs is to ask writers to review the responses of the other three writers.
Because the writer has already read the story, they do not need to fully re-read the story, and because they have answered the questions previously, they already have a sense of what constitutes a good response to each question.

In each validation task, we show the reviewer the original story, the set of five questions, and three responses for each question written by other writers. 
Reviewers are first asked to annotate spans of the responses that contain typos or factual errors.
Next, they are asked to rank the three responses from best to worst.
We instruct the reviewers to rank the responses by (1) how well the response correctly answers the question; (2) how well the summary includes all relevant details; (3) how well the response draws from multiple parts of the story, using their judgment to balance the three factors.
Writers are informed during the writing phase that their responses will be evaluated along these dimensions.
Finally, reviewers are tasked with providing written feedback for each response about how that response could be improved.
The feedback is provided to writers between batches of work to help them improve their responses.
Reviewers report that this phase typically takes 20--30 minutes.

Afterwards, for each question, we compile the individual reviewer rankings into an aggregate ranking.
We incentivize high-quality writing by awarding bonus payments to writers based on their response's placement in the overall ranking.
We pay \$2.50, \$1.25, \$0.75, \$0.50 for ranking first, second, third, and fourth respectively.\footnote{In case of ties, we sum the bonuses for the tied positions and distribute them evenly.}
The average bonus is \$1.25 per response, so writers earn an average additional bonus of \$6.25 per story.
Workers are informed of the bonus structure before writing.

Similarly, we incentivize high-quality reviewing by awarding bonus payments to reviewers based on how well their rankings agree with the aggregate ranking.
For each pair of responses, we pay a reviewer a bonus of \$0.50 if their ranking of the pair agrees with the aggregate ranking (i.e., if both the aggregate and reviewer's ranking say response $A$ > response $B$), so reviewers can earn up to \$1.50 per question and \$7.50 per story.
On average, individual reviewers agree with the aggregate ranking on pairwise comparisons 76\% of the time, corresponding to an average bonus of \$5.57 per story.

\subsection{Writer Details}

Because our tasks are very time-consuming and detail-oriented, we eschew crowdsourcing platforms like Amazon Mechanical Turk where eliciting high-quality responses for these types of tasks can be challenging.
Instead, we hire a small group of skilled writers for long-term contracts, drawing both from Upwork\footnote{\url{https://www.upwork.com/}} freelancers and undergraduate students from our institution.
Specifically, we hire 11 Upwork writers and 7 undergraduates.\footnote{We use two worker populations due to spending limits on Upwork. The two populations are not mixed, i.e. undergraduates do not review Upwork writers' responses and vice versa.}
We discuss some qualitative differences between the two populations in Appendix~\ref{ax:crowdsourcing_comparison}.
Most writers create 20--40 responses for the dataset, although five authors submitted 10 or fewer responses.
All writers are informed that their writing will be released publicly for use in AI development.

Our Upwork writers are typically US-based native English speakers. 
Many of them are college-educated, frequently with degrees in the humanities and prior experience in professional copywriting and editing.
We found workers for our task by posting an open call on Upwork to participate in a paid interview.
In the interview, applicants review an example writing task with sample questions and responses, and then complete a practice writing task.
We hired the top 33\% of writers based on their performance on the interview task after manually reviewing their responses.
We pay Upwork workers \$13 and \$8 for each writing and reviewing task respectively, with additional opportunities for bonuses described above. 

The undergraduates we hire are all English-fluent and come from diverse nationalities and areas of study---the smaller and more junior pool of applicants prevents us from focusing as much on relevant experience as we do with Upwork.
Students are paid a constant \$20/hr.\footnote{Due to the structure of student employment contracts, we are unable to pay students using the bonus payment structure and we instead periodically manually review their responses to ensure they are high-quality.}
Students are hired based on relevant experience and writing samples.
After they are hired, we show them the same example task and have them do the practice writing task that we showed the Upwork workers.


\begin{table*}[t]
    \centering
    \begin{tabular}{llrrrcc}
        \toprule
        Dataset & Domain & \# Examples & Doc. Len & Summ. Len & Multi-ref? & Public? \\
        \midrule
        CNN/DM & news & 311K & 804 & 60 & \redxmark & \redxmark \\ 
        XSum & news & 226K & 438 & 24 & \redxmark & \redxmark \\
        BookSum & fiction, Sparknotes & 12K & 5102 & 505 & \redxmark & \redxmark \\
        QMSum & meeting transcripts & 1808 & 9067 & 70 & \redxmark & \greencheck \\
        SQuALITY & sci-fi stories & 625 & 5200 & 237 & \greencheck & \greencheck \\
        \bottomrule
    \end{tabular}
    \caption{Summary statistics for various summarization datasets. For BookSum, we consider the chapter-level version. The number of examples is across all splits. For question-based summarization datasets (SQuALITY and QMSum) we count examples as number of unique document-question pairs. Statistics for datasets are borrowed from original dataset papers; statistics for CNN/DM and XSum were borrowed from \citet{kryscinski2021booksum}. CNN/DM and XSum are often available online in practice, but distributing the dataset is legally questionable.}
    \label{tab:datasets}
\end{table*}

\section{SQuALITY}

We present summary statistics of SQuALITY and other summarization datasets in Table~\ref{tab:datasets} and examples from the dataset in Table~\ref{tab:dataset_examples_short}.
SQuALITY is available at \url{https://github.com/nyu-mll/SQuALITY} under a CC BY license.

\subsection{Data Size and Splits}

SQuALITY consists of 100 stories that are split 39/25/36 across the train/validation/test splits (or, equivalently, 195/125/180 document-question pairs).
SQuALITY contains a similar number of summaries to QMSum \citep{zhong-etal-2021-qmsum}, another crowdsourced summarization dataset, but SQuALITY contains four references per example and thus fewer input documents.
This difference in allocation arises from the crowdsourcing protocol: In creating SQuALITY, we have writers review each other's work while in creating QMSum, the authors manually review all responses.
Protocols wherein workers review each other work are more scalable.
We also argue that the presence of multiple references per input is useful for model evaluation, as automatic metrics such as ROUGE were originally developed on multi-reference datasets.
While naive multi-reference ROUGE still correlates poorly with human judgments of quality for SQuALITY (see Section~\ref{sec:multi_ref}), having a diverse set of references opens up opportunities for the development of new evaluation metrics that take into account the diversity of acceptable summaries for a given input, even in the question-focused setting.

Additionally, we assign stories to splits to be consistent with the QuALITY dataset \citep{pang2021quality}, such that stories that appear in both QuALITY and SQuALITY are assigned to the same split.
We leave the exploration of hybrid summarization-QA models and extrinsic QA-based summarization evaluation to future work.

\subsection{Length}

Documents are an average of 5199.4 tokens long without punctuation (standard deviation 522.4 tokens).\footnote{We use the \texttt{en\_core\_web\_sm} spaCy tokenizer.}
The minimum document length is 3473 tokens and the maximum is 6165 tokens.
The documents in SQuALITY are very close in length to the documents in the chapter version of BookSum, which consists of chapters of public domain novels.
The input documents are shorter than the meeting transcripts of QMSum, which, being dialogue, contain more disfluencies and off-topic utterances. 

Questions are 8.9 tokens on average with a minimum length of 6 tokens and a maximum of 12 tokens.
Responses are 237.1 tokens long on average (standard deviation 132.5).
The plot summaries have an average length of 441.9 tokens (standard deviation 90.9 tokens) and are comparable in length to those of BookSum.
The responses to the other questions are shorter with an average length of 185.9 tokens (standard deviation 82.4 tokens), but are still longer than the summaries in QMSum.


\begin{table}[t]
    \centering
    \begin{tabular}{lrrrr}
    \toprule
    \multirow{2}{*}{Text} & \multicolumn{4}{c}{N-gram Size} \\
    & 1 & 2 & 3 & 4 \\
    \midrule
    Random & 19.7 & 2.7	& 0.1 & 0.0 \\
    Same story & 27.4 & 5.8 & 1.2 & 0.4 \\
    Same question & 33.4 & 8.7 & 2.3 & 0.8 \\
    \midrule
    \midrule
    Story & 69.4 & 22.0 & 5.0 & 1.7 \\
    
    \bottomrule
    \end{tabular}
    \caption{(Top) Average percentage of unique n-grams shared between pairs of responses from different sources: two different stories, different questions but the same story, and the same question. (Bottom) Average percentage of unique n-grams are shared between a response and the corresponding story.}
    \label{tab:overlap}
\end{table}

\subsection{Response Diversity}

We verify that the summaries are abstractive by computing the percentage of response n-grams that also appear in the input story, which we show in Table~\ref{tab:overlap}.
The high recall of 1-grams is unsurprising given the long length of the stories, but the low recall of 3- and 4-grams indicate that the responses are highly abstractive, which makes sense given that the responses need to compress the stories by 95.4\% on average. 

We next consider the diversity between pairs of responses to the same question.
If responses are similar, then collecting multiple references is potentially wasteful.
We show the average percentage of unique n-grams shared between responses to the same question in Table~\ref{tab:overlap}.
The overlap is quite low: 33\% of unigrams (around 75 tokens for the average 237 length response) and less than 10\% of bigrams. 
This overlap is only slightly higher than the average overlap between responses to completely different stories.
The wide range of responses to the same question highlights how diverse the summarization task is, a fact that is made evident in SQuALITY but not in single-reference datasets.

\begin{table*}[t]
    \centering
    \begin{tabular}{lrrrrrr}
    \toprule
    Model & \# Params & ROUGE-1 & ROUGE-2 & ROUGE-L & METEOR & BERTScore \\
    \midrule
    LED & 160M & 27.7 & 5.9 & 17.7 & 16.5 & 82.7 \\
    PEGASUS & 540M & 38.2 & 9.0 & 20.2 & 23.4 & 84.9 \\ 
    BART & 340M & 40.2 & 10.4 & 20.8 & 24.5 & 85.3 \\ 
    BART+DPR & 340M & \textbf{41.5} & \textbf{11.4} & \textbf{21.0} & \textbf{26.1} & \textbf{85.5} \\ 
    \midrule
    Human$^*$ & - & 46.6 & 12.5 & 22.7 & 30.6 & 86.2 \\
    \bottomrule
    \end{tabular}
    \caption{Automatic evaluation results. LED and PEGASUS summaries tend to be a single sentence repeated, and this is reflected in low metric scores. BART and BART+DPR perform better, though human evaluation demonstrates that the gap between human-written summaries and BART-written summaries is much larger than indicated by automatic evaluation metrics (see Table~\ref{tab:human_eval}). $^*$The human reference is evaluated against three other references while the model-generated summaries are evaluated four references, artificially raising their score.}
    \label{tab:automatic_eval}
\end{table*}
\begin{table}[t]
    \centering
    \begin{tabular}{lrrr}
    \toprule
    Model & Corr. & Coverage & Overall \\
    \midrule
    BART & 34.8 & 15.6 & 18.1 \\
    BART+DPR & 45.4 & 24.3 & 27.9 \\
    Human & \textbf{94.1} & \textbf{88.8} & \textbf{91.3} \\
    \bottomrule
    \end{tabular}
    \caption{Human evaluation results for two models and a human-written response. Corr. stands for correctness. Ratings for each property are averaged across 3 workers, then averaged across questions.}
    \label{tab:human_eval}
\end{table}

\section{Baselines}

\subsection{Models}

For our baselines, we evaluate supervised sequence-to-sequence models using different pretrained language models as the base model.
We do not explore prompting approaches for summarization with closed-access models.
Previous work has found that zero-shot prompting of models to summarize can produce high-quality summaries \citep{radford2019language,wu2021recursively}, though public models like GPT-3 do not have the capacity to process full stories from our dataset.
We implement our baselines using the pretrained models available via HuggingFace Transformers \citep{wolf-etal-2020-transformers}.

\paragraph{BART} BART \citep{lewis-etal-2020-bart} is a Transformer-based \citep{vaswani2017attention} encoder-decoder model pretrained on a token in-filling objective and a sentence permutation objective.
We use \texttt{BART-large}, which has a maximum input sequence length of 1024 tokens, so we truncate stories dramatically to fit this simple baseline.

\paragraph{BART+DPR} We experiment with an extract-then-summarize baseline. 
Instead of truncating stories when using BART, we retrieve story sentences that are most relevant to the question and concatenate them to form the input.
Specifically, we use the pretrained Dense Passage Retriever \citep{karpukhin-etal-2020-dense} that encodes the question into a vector representation and retrieves the story sentences that are most similar to the question.

\paragraph{PEGASUS} PEGASUS \citep{zhang2020pegasus} is a Transformer-based encoder-decoder model that is pretrained using an objective designed especially for summarization. 
Specifically, it is pretrained to predict masked out sentences rather than masked out words. 
The masked sentences are selected to be pseudo-summaries of the overall document.
PEGASUS is pretrained on sequences of at most length 512, but we follow previous work in finetuning \texttt{PEGASUS-large} with a max sequence length of 2048 tokens, truncating stories to fit.
    
\paragraph{LED} Longformer Encoder-Decoder \citep{beltagy2020longformer} is an encoder-decoder model where the encoder is a Longformer and the decoder is a Transformer.
A Longformer modifies the Transformer architecture with a more efficient self-attention pattern that allows the model to efficiently scale to long documents. 
Specifically, LED has a maximum input sequence length long enough to fit the entire story. We use a context length of 8192 for memory efficiency.
LED is semi-pretrained: It initializes its parameters using the weights of BART, copied eight times over (since the LED context length is eight times that of BART).
We use \texttt{LED-base}.

\subsection{Training}

We format example inputs by concatenating the question to the beginning \textit{and} end of the document, separated by a special \texttt{[SEP]} token, based on previous work on question-focused summarization \citep{vig2021exploring}.
Each (story, question, reference) tuple is mapped to a separate training instance, so each (story, question) input is associated with four training examples, one per reference.
We finetune models using the AdamW optimizer \cite{loshchilov2018decoupled}.
At test time, we generate summaries using beam search with beam width 4.

\subsection{Evaluation}

We evaluate our baselines with ROUGE \citep{lin-2004-rouge} and METEOR \citep{banerjee-lavie-2005-meteor}, standard automatic metrics for summarization.
We also evaluate with BERTScore \citep{zhang2019bertscore}, which uses BERT to compute the similarity between references and model generations.\footnote{We use a version of BERTScore based on \texttt{RoBERTa-large.}}
For all metrics, we report F1.
All automatic metrics used handle multiple references by evaluating a candidate against each reference individually, and then taking the max score across references.
We assume that these automatic evaluation metrics used are flawed and recommend human evaluation as a gold standard, which we describe in Section~\ref{sec:human_eval}.


\subsection{Automatic Evaluation Results}
\label{sec:auto_eval}

We present results using various automatic evaluation metrics in Table~\ref{tab:automatic_eval} and examples of model generations in Table~\ref{tab:model_outputs}.
We observe that LED fails to learn the task and generally produces outputs containing long, repeated sentences.
The pathological behavior is reflected in the low ROUGE-1 and ROUGE-2 scores for the model.
We hypothesized that the poor performance is because the small dataset size is not enough to finetune the additional positional embeddings.
We additionally explored transfer learning approaches where the model was first finetuned on a larger long-context summarization dataset, such as arXiv \citep{cohan-etal-2018-discourse} or GovReport \citep{huang-etal-2021-efficient}, and then finetuned on SQuALITY.
However, training on intermediate datasets did not fix the issue of degenerate outputs, indicating that the additional positional embeddings were not the bottleneck in the models' performance on SQuALITY.
Overall, we found that public pretrained models for medium to long input tasks were not effective off the shelf.

PEGASUS, BART, and BART+DPR do substantially better on the task and produce sensible outputs, despite having partial inputs.
PEGASUS slightly underperforms the BART variant according to the metrics.
BART+DPR outperforms BART with truncated input across all metrics. 

Additionally, we evaluate the human references using the automatic metrics by holding one reference out and comparing it with the various metric against the remaining three references. We repeat this process for all references and average the metric score across held-out references. While this use of three references rather than four disadvantages the human references (see Section~\ref{sec:multi_ref}), we still find that they score higher than machine outputs.

\begin{table*}[t]
    \small
    \centering
    \begin{tabularx}{\textwidth}{XX}
    \toprule
    \multicolumn{2}{l}{Title: \href{https://www.gutenberg.org/ebooks/61146}{Retief of the Red-Tape Mountain} (\url{https://www.gutenberg.org/ebooks/61146})} \\ 
    \midrule
    \multicolumn{2}{l}{Q: What is the relationship between the Jaqs and the Terrestrials throughout the story?} \\ 
    \midrule
    \textbf{Reference}: The Jaqs and the Terrestrials fight each other throughout the story. It started when a human saw a Jaq and thought it was some type of native game and shot it. From that incident, the Jaqs concluded that the humans were sportsmen like themselves and responded by going to one of the farms and killing two cows. Since then, the two sides have been attacking back and forth, and the humans think the Jaqs are fighting against them... 
    &
    \textbf{LED}: Retief is a vice-consul at the Embassy of the Mosaic of the Two Dawns. He is in charge of the affairs of the Embassy and is responsible for keeping the diplomatic relations between the two planets in check. He is also responsible for keeping the diplomatic relations in check by sending Retief on his expeditious trip to the planet Adobe. When Retief arrives at the planet, he is greeted by a large Flap-jack, a creature with talons that look like lobsters... 
    \\
    \midrule
    \textbf{BART}: The Terrestrials and the Jaqs have a tense relationship throughout the story. The Terrans have attempted to establish contact with the native life form, the Jaq, in order to try to gain their trust and gain information about their native life forms. The Jaqs are hostile to the Terrans because they consider them to be an invasive species that are trying to take over their home planet, which they consider to be uninhabited. The Jaqs have a history of war with the Terran settlers... 
    & 
    \textbf{BART+DPR}: The Terrestrials and the Flap-jacks are an alien race that live on the planet Adoban. They are hostile to humans and have attempted to stir up trouble with an intelligent alien life form, the Jaq, three months ago. The humans are attempting to establish trade with the aliens in order to gain access to the planet’s resources, but the aliens are having none of it. They have no intention of trading with the humans and are only interested in trading with them for food and... 
    \\
    \bottomrule
    \end{tabularx}
    \caption{Example model generations on \dataset.}
    \label{tab:model_outputs}
\end{table*}

\section{Human Evaluation}
\label{sec:human_eval}

Automatic metrics for evaluating text summarization have been well-documented as correlating poorly with various human judgments of quality \citep{schluter-2017-limits,kryscinski-etal-2019-neural,durmus-etal-2020-feqa}.
As such, we accompany automatic evaluation of the baseline systems with human evaluation.
Specifically, we ask Upwork workers to rate the quality of outputs from BART and BART+DPR generated with beam search on the test data.

For each task, we show the worker a story from the test set and the five questions for that story.
For each of the questions, we show the two model-generated summaries and a human reference.
As the task is labor-intensive, we use four of the same Upwork writers for the human evaluation as for the data collection phase. 
Workers may have previously read the story and thus answered the questions, and we are careful to not show workers their own responses.
If they have not previously read the stories, workers are paid to read the story.
Workers are informed that the responses are a mixture of human- and machine-written, but not informed which responses are which.
We pay workers \$8/task and an additional \$8 if they have not previously read the story.
All workers complete the same number of tasks.

For each response, we ask workers to rate the response for three properties: correctness, coverage, and overall quality.
The evaluation UI and property definitions are available in Appendix~\ref{ax:human_eval}.
For each property, the response is rated on a scale from 1-100, similar to direct assessment ratings in machine translation \citep{bojar-etal-2016-results}.
We instruct workers to assign ratings that align with their preference rankings between systems, similar to \citet{sakaguchi-van-durme-2018-efficient}.
We annotate 20 stories (100 questions) this way, with three Upwork workers completing each task.
For each property, we average the ratings across annotators. 

We present results of the human evaluation in Table~\ref{tab:human_eval}. 
The standard deviations of property ratings across questions are shown in Table~\ref{tab:human_eval_std} in Appendix~\ref{ax:human_eval}.
For all questions and all properties, all human annotators rank the human-written response as better than the model responses.
The human-written response has an average rating around or above 90 for all three properties.
On the other hand, BART and BART+DPR have an average rating below 50 for all three properties, substantially below corresponding ratings for the human response.
Across all three properties, BART+DPR is ranked as better than BART on 70\% of examples.
The models receive the highest rating on the correctness property among all properties.
Upon inspecting the model generations, we partly attribute these relatively high ratings to the fact that the model-generated responses are fairly generic and devoid of specific details.
This lack of specificity is reflected in the especially low coverage ratings of the model-generated summaries.
Overall, we conclude that fully-public automatic summarization systems still lag significantly behind human writers.


\begin{table}[t]
    \centering
    \begin{tabular}{lrrr} 
    \toprule
    Metric & \makecell[r]{Model \\ Only} & \makecell[r]{Human \\ Only} & All \\
    \midrule
    ROUGE-1 & -7.4 & 6.8 & 63.1$^*$  \\
    ROUGE-2 & -7.8 & 5.2 & 42.8$^*$  \\
    ROUGE-L & -3.2 & 14.0 & 47.8$^*$ \\
    METEOR & -11.1 & -4.3 & 54.1$^*$ \\
    BERTScore & 5.5 & 0.8 & 68.7$^*$ \\
    \bottomrule
    \end{tabular}
    \caption{Pearson correlation between automatic evaluation metrics and human judgments of overall quality. 
    Correlations are near zero and not statistically significant when only considering model-generated summaries (`model only') or only human-written summaries (`human only'). 
    Correlations are significantly positive ($^*$) when considering human- and model-written summaries together (`all').
    } 
    \label{tab:correlations}
\end{table}

\subsection{Correlation Between Automatic and Human Evaluation}
\label{sec:metric_corr}

We next consider how well the automatic metrics used correlate with the collected human judgments by computing the Pearson correlations between the two.
We consider the correlations for three subsets of the collected data: only model-written summaries (200 summaries), only human-written summaries (100 summaries), and all summaries (300 summaries).
We present the correlations with the judgments of overall summary quality for these subsets in Table~\ref{tab:correlations}.
Correlations with human judgments of other properties are generally similar.

When considering model-written and human-written summaries together, all metrics have a substantial positive correlation with the human judgments of overall summary quality.
These positive correlations reflect the fact that the automatic metrics rank human-written summaries as better than model-written ones, which is consistent with human evaluators, though the magnitudes of differences between model- and human-written summaries are substantially different.

However, when considering only model-written summaries or only human-written summaries, the correlations are substantially weaker, even slightly negative in the case of only model-written summaries (no correlations are significant in these settings).
The weak correlations in these settings point to the brittleness of using these automatic metrics when comparing the outputs of two automatic summarization systems, where metric values will similarly be in a narrow range.
Additionally, given that automatic summarization models only slightly trail behind human responses (see Table~\ref{tab:automatic_eval}) but human evaluators rate them as substantially different, we argue that existing automatic evaluation metrics such as ROUGE do not adequately reflect the differences between model- and human-written summaries.
In light of these findings, we caution against relying on automatic evaluation metrics to measure system quality on SQuALITY and instead rely on human evaluation of model outputs.

\begin{table}[t]
    \centering
    \begin{tabular}{lrrr}
    \toprule
    Metric & Avg. & Max. & $\Delta$ \\
    \midrule
    ROUGE-1 & 37.9 & 41.5 & 3.6 \\
    ROUGE-2 & 8.7 & 11.4 & 2.1 \\
    ROUGE-L & 18.8 & 21.0 & 2.2 \\
    METEOR & 22.7 & 26.1 & 3.4 \\
    BERTScore & 84.8 & 85.5 & 0.7 \\
    \bottomrule
    \end{tabular}
    \caption{Average and maximum metric value across the four references for BART+DPR.}
    \label{tab:avg_max_metric}
\end{table}

\subsection{Automatic Metrics with Multiple References}
\label{sec:multi_ref}

We next consider whether having multiple references improves the correlation of automatic evaluation metrics. 
Standard automatic evaluation metrics for summarization like ROUGE were originally developed on multi-reference datasets.
Recent summarization datasets are predominantly single reference, and this mismatch may contribute to the poor correlation of ROUGE with human judgments of quality for these datasets \citep[][i.a.]{pang-etal-2021-agreesum,pagnoni-etal-2021-understanding,scialom-etal-2021-questeval}.
Because SQuALITY is multi-reference, we can use the dataset to measure the effect of varying the number of references used in automatic metrics on the correlation with human judgments.

We find that using fewer references when computing the automatic evaluation metrics does not substantially change the correlations with human judgments.
To demonstrate why, we show the average and maximum metric values for each automatic metric in  Table~\ref{tab:avg_max_metric}.
We observe that for all metrics considered, the maximum value of the metric is relatively close to the average metric value across references.
In other words, despite having diverse references, the metric values are similar across references.
Thus, using multiple references does not improve correlations between automatic metrics and human judgments of overall quality.
However, we note that simply taking the maximum metric value over references is relatively simple, and that there may be more sophisticated ways to use the diverse references to compute generation quality.



\section{Conclusion}

We present SQuALITY, a high-quality, long-context dataset for question-focused summarization.
Because SQuALITY summaries are crowdsourced rather than found, we can choose input documents that are of an accessible domain and under an open license.
Our crowdsourcing protocol allows for multiple summaries and references per input while making the cost of data collection more tractable.

Baseline results with competitive public medium-scale pretrained models suggest that the dataset remains beyond the capabilities of such systems.
Our best performing model is an extract-then-summarize model where we use the input questions to retrieve story sentences as input. The performance of proprietary larger-scale models remains an open question, and may depend significantly on whether such models can process the full length of SQuALITY examples without truncation.

Given the poor correlation of existing automatic metrics with human judgments of model outputs, we expect that automatic metrics will provide a very weak signal for progress on SQuALITY. 
We recommend that researchers using SQuALITY evaluate their summarization systems by having human annotators read a selection of our source stories and compare model outputs on those stories.
To facilitate this, we will make our templates for human evaluation available, though creating efficient and effective methods for evaluating summaries of long input documents remains an open issue.

\section*{Ethical Considerations}

We expect this work to advance two outcomes: (i) accelerated progress in language modeling, especially toward controllable text generation and long-text comprehension, and (ii) an increase in the hiring of professional and/or crowdworker writers by researchers and product developers in this area. Both of these have potentially significant costs and benefits that are beyond the scope of this paper to investigate.

More concretely, the stories in the dataset were written between 1930--1970 and therefore contain dated and potentially harmful stances on topics like race and gender.
Models trained on the data may reproduce these stances, especially if they are trained on the complete texts, rather than the reference summaries alone. We are releasing SQuALITY primarily for use as a research benchmark, and we recommend extreme caution if SQuALITY is used as part of the training set for any deployed system.

Further, the summaries in the dataset were created by writers that are primarily college-educated and either native-English or English-fluent.
A system that does well on our dataset only demonstrates competence in mainstream US English, and may not generalize to other variants of English. 

\section*{Acknowledgements}

We thank the writers who created and reviewed the summaries:
Christina Li,      
Cathy Liu,         
Mateo Pardo,       
Alexandra Rumyantseva, 
Pei-Ling Wu,        
Alicia Chatten,    
Dolly Farha,           
Jamie Swanger,     
Isaiah Swanson,    
and other anonymous writers.

We also thank the members of the ML$^2$ lab at NYU for providing helpful feedback in the early stages of this project, particularly Nitish Joshi, Nikita Nangia, and Kyunghyun Cho.
Finally, we thank Peter Liu, Wojciech Kryściński, Sebastian Gehrmann for helpful discussions about summarization datasets.

This project has benefited from financial support to SB by Eric and Wendy Schmidt (made by recommendation of the Schmidt Futures program) and Apple, and from in-kind support by the NYU High-Performance Computing Center and Google Cloud.
This material is based upon work supported by the National Science Foundation under Grant Nos. 1922658 and 2046556.
Any opinions, findings, and conclusions or recommendations expressed in this material are those of the author(s) and do not necessarily reflect the views of the National Science Foundation.

\bibliography{anthology,custom}
\bibliographystyle{acl_natbib}


\appendix

\section{Crowdsourcing Details}
\label{ax:crowdsourcing}

\subsection{Question Templates}
\label{ax:question-templates}

We provide the following question templates to the writers: 
\begin{itemize}
    \item What is the plot of the story?
    \item What happens to [character X] throughout the story?
    \item What is the relationship between [character X] and [character Y]?
    \item What is the setting of the story?
    \item What is the significance of [object X] on the rest of the story?
    \item How is [theme X] explored throughout the story?
    \item Story-specific questions
\end{itemize}

Writers always answer the question ``What is the plot of the story?''. For more subjective templates such as ``What is the significance of [object X]?'' or ``How is [theme X] explored?'', we ask the writers to use these templates only in cases where they believe the answer will be clear and unambiguous to someone who has read the story carefully.

\subsection{Crowdsourcing Interfaces}

\begin{figure*}[ht!]
     \centering
     \includegraphics[width=0.99\textwidth]{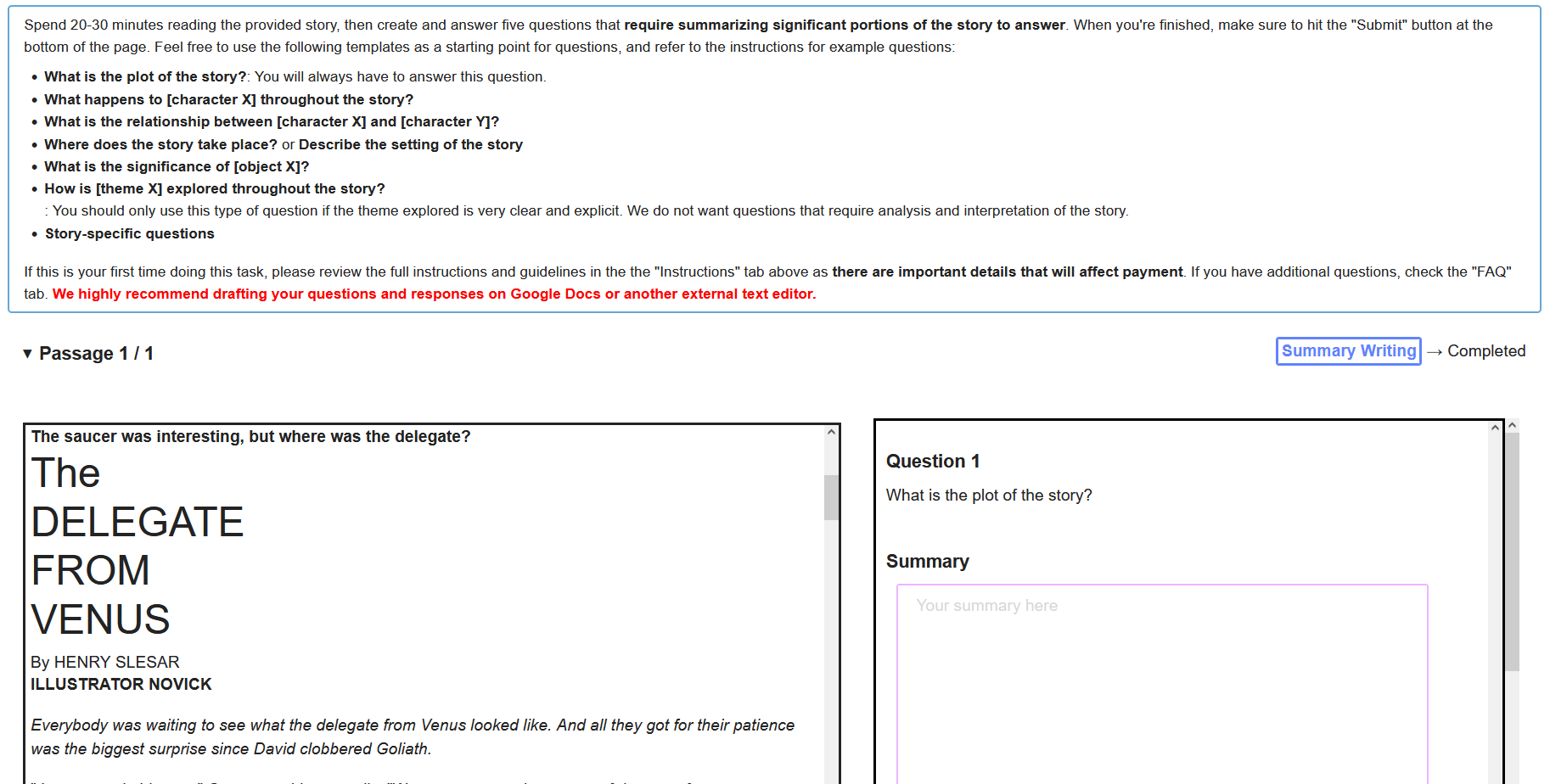}
    \caption{Screenshot of the writing UI. Workers are shown the story on the left and five questions on the right, and they are tasked with writing responses to each of the questions. If the worker is the first person to work on a story, they write four questions about the story to answer (The question ``What is the plot?'' is always asked), and we provide the worker with a list of question templates in the UI to help them write good questions.}
    \label{fig:writing_ui}
    \vspace{-0.3cm}
\end{figure*}

\begin{figure*}[ht!]
     \centering
     \includegraphics[width=0.99\textwidth]{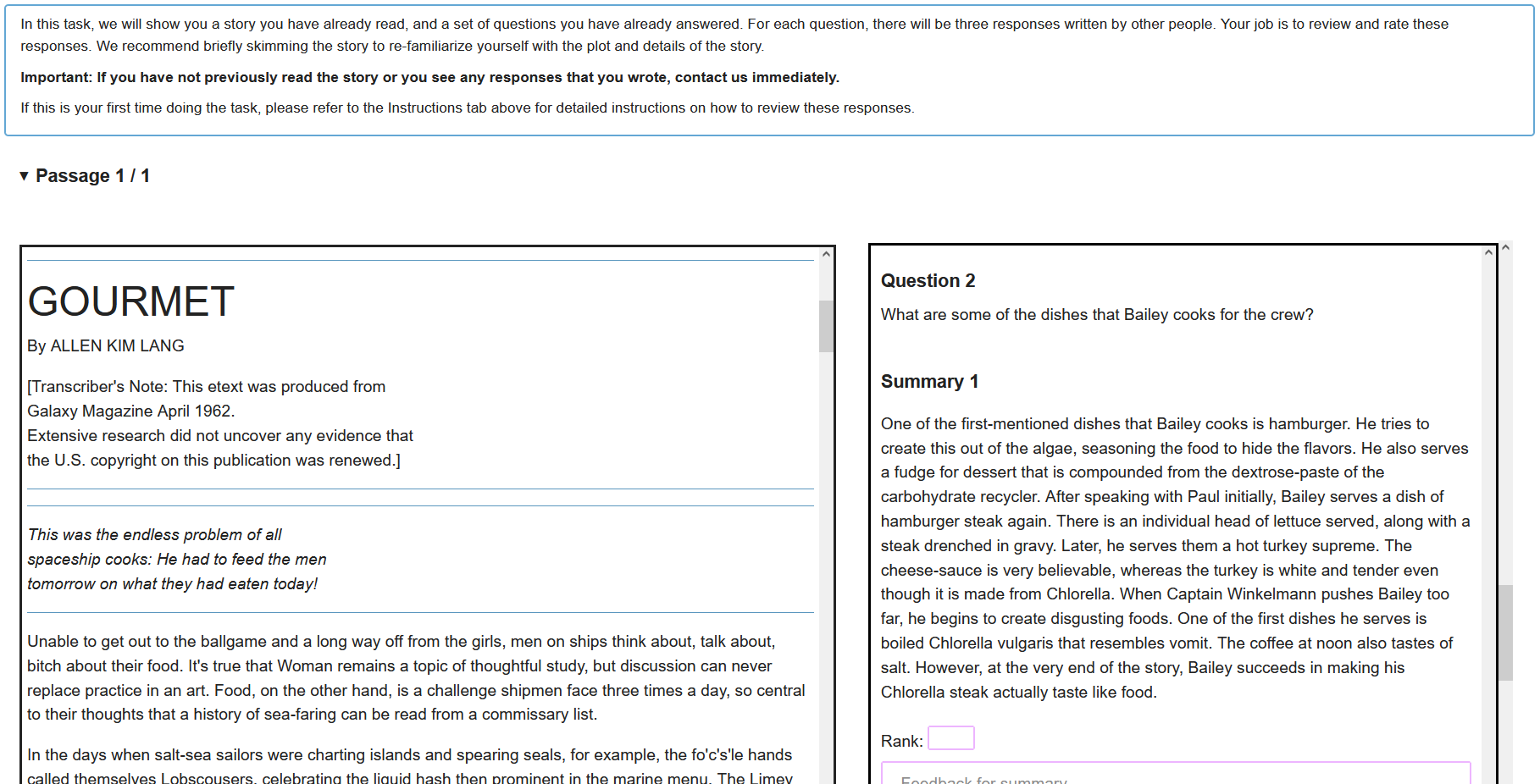}
    \caption{Screenshot of the reviewing UI. Workers are shown the story on the left and five questions on the right. Each of the questions has three responses that the worker is tasked with ranking from best to worst. Additionally, for each response, the worker is instructed to highlight typos and factual errors, as well as provide written feedback to the writer. This feedback is later provided to the writer to help them improve their responses in subsequent rounds of writing.}
    \label{fig:reviewing_ui}
    \vspace{-0.3cm}
\end{figure*}

We show screenshots of our UIs and abbreviated task instructions for writing and reviewing summaries in Figures~\ref{fig:writing_ui} and \ref{fig:reviewing_ui}, respectively.

\subsection{Comparing Upwork and Undergraduates}
\label{ax:crowdsourcing_comparison}

Generally, we found that both Upwork and undergraduate workers took the task seriously and produced quality summaries.
Writers from Upwork qualitatively produced slightly higher quality responses, perhaps because we were able to filter more aggressively for relevant backgrounds and skills when hiring on Upwork.
Hiring writers on Upwork was more expensive than hiring student writers.

Anecdotally, the workers we hired from both populations enjoyed the tasks, and we see this as a significant advantage to using popular fiction in benchmark tasks.
However, we did find that some Upwork contractors quit our task during the course of data collection, and some mentioned that our task paid less than other tasks on Upwork.
Because students were hired for long-term contracts (on the order of months), they did not drop out of the data collection process, but working with them did require careful work scheduling around exams and breaks.

\section{Dataset Examples}
\label{ax:dataset_examples}

\begin{table*}[t]
\small
    \centering
    \begin{tabularx}{\textwidth}{XX}
    \toprule
    \multicolumn{2}{l}{Title: \href{https://www.gutenberg.org/ebooks/51656}{Pick A Crime} (\url{https://www.gutenberg.org/ebooks/51656})} \\ 
    \midrule
    \multicolumn{2}{l}{Q: What is the CPA and what does it do?} \\ 
    \midrule
    The Crime Prevention Association is an organization that stops crime. Instead of capturing criminals, the goal of the Association is to prevent the crime from ever happening. They implement thousands of crime-prevention methods and devices. There are many amateur cops who constantly follow criminals around in hopes of catching them in the act so that they may be hailed a hero and given a promotion. Hendricks even explains that the kids have junior CPA clubs, where they record the criminals in little cardboard boxes. They will also follow the criminals around until they die. There are millions of microphones hidden by the CPA everywhere, and any threatening messages are sent to the CPA Brain. The CPA Brain is a monster electronic calculator that can alert police helicopters of any threatening messages, and there are also many hidden TVs and metal detectors. For Arson, heat detectors exist too, and chemical poisoning has made it impossible for people to get poisoned. There are shock treatments, encephalographic devices, a form of prefrontal lobotomy, and a dozen other treatments to reform criminals. 
    & 
    The CPA, Crime Prevention Association, is a system that detects different kinds of crimes and prevents them from happening. Thousands of robots and devices make crimes impossible. The association will not punish any crime, instead, the criminal will be send to a CPA hospital for some treatments that will result in getting the best jobs. The CPA also hands out ID cards that states one’s tendency to commit crimes. The CPA has robot bartenders that can detect the drunkenness of a person and prevent anyone from actually getting drunk. There is WSDA teaching judo and jujitsu to women. There are spy cameras and speakers in each alley and street watching every person all the time to prevent all kinds of crimes. The CPA Brain will catch sentences that indicate crimes and watch them more carefully. There are heat-detectors, gun and knife detector, chemical detectors, etc. The CPA brainwashes people, making them believe that crimes are filthy. The treatment will make the criminal’s brain catch every attempt that he or she tries to commit a crime and prevents it from happening. 
    \\
    \midrule
    The CPA is Crime Prevention Organization. It fights crime by all means and reduces its rates to a very small level. They put microphones and detectors everywhere to hear the conspiracies. They place robots as bartenders to control the level of alcohol in visitors to prevent them being drunk. They make all the women learn self-defense. The organization's made crime almost impossible and they do not punish for it, but prevent. All who tried to commit a crime are given free treatment. The CPA hospitals treat those few criminals for free and make them unable to commit any further crime. CPA seems to be everywhere, those who tell about the crime are highly rewarded. Neon signs, TV, radio and other means constantly remind people that crime is filth. 
    & 
    The CPA is meant to prevent crime and not punish crime. It stands for Crime Prevention Association. The CPA organization has made crime nearly impossible through various methods of surveillance and intelligence gathering. The crime was not punished by the CPA but addressed by sending the person to a hospital for expensive treatment to correct and remove the deviance from the person’s mind. A CPA ID card is required to be carried by everyone and when asked, a person has to present the ID card. Being drunk is illegal according to the rules of the CPA.
    \\
    \bottomrule
    \end{tabularx}
    \caption{The four full human-written references from Table~\ref{tab:dataset_examples_short}.}
    \label{tab:dataset_examples_long}
\end{table*}

Table~\ref{tab:dataset_examples_long} shows the full references for the example in Table~\ref{tab:dataset_examples_short}. 
Table~\ref{tab:dataset_examples_additional} shows additional examples from SQuALITY.

\begin{table*}[t]
\small
    \centering
    \begin{tabularx}{\textwidth}{XX}
    \toprule
    \multicolumn{2}{l}{Title: \href{https://www.gutenberg.org/ebooks/61053}{Tolliver's Orbit} (\url{https://www.gutenberg.org/ebooks/61053})} \\
    \midrule
    \multicolumn{2}{l}{Q: Describe the equipment used throughout the story.} \\ 
    \midrule
    Tolliver is a pilot, but while at the Ganymede branch he drives a tractor. One of the equipment used during the story is the automatic flight. An automatic flight allows loaded ships to take a slow and economical orbit using automatic signaling equipment towards Earth. As the loaded ship gets closer to Earth, it is boarded by pilots that land the ship. Another piece of equipment mentioned are spacesuits. The spacesuits involve valves and seals and microphones for people to communicate with each other in the spacesuits. The communication is activated by a switch under the chin on the helmet of the spacesuit. They also come with a heavy knife.
    &
    Various types of transportation are used throughout the story - tractors to travel on Ganymede between the city and the spaceport, spaceships requiring a lot of fuel and economy orbits which require less fuel but take much longer to get to the place. In a storeroom there are plenty spacesuits, some of which need replacement. Knives are standard suit equipment. Spaceships are equipped with airlocks, ladders and switch-cover. In the control room there is an acceleration seat, a button to set off, a radio and TV, with a screen to see the other side of the call. 
    \\
    \midrule
    Tolliver is first assigned to use an airtight tractor to transport to and from the spaceport. This tractor is like a regular one, but built specifically to trek across Ganymede with its gravity. When Tolliver and Betty are locked into Jeffers' office, he uses a lighter and paper to bend the plastic of the door. Then, he uses a knife to cut through the plastic of the dome. Finally, Tolliver and Betty board a ship, where the orbit is automatically preset in order to preserve fuel. The ship, which Tolliver knows how to operate, is airlocked. Betty uses a transmitter to contact Space Patrol.
    &
    Firstly, Tolliver takes Betty towards Jeffers’ office on a tractor since it can go through the frozen surface of Ganymede. Then later, when Betty and Tolliver were put in the empty office, Tolliver uses a lighter to light up the mess of discarded records so that the plastic can be bent. Later, inside the storage room, Tolliver finds some spacesuits for the two to wear. Then finally, when they gets to the control room, they gets onto the acceleration seat. Using the ship, the two fly into the economy orbit for Earth in order to escape. In the end, Betty uses the scanner and microphone to make a call to the Space Patrol so that they will arrest Jeffers.
    \\
    \midrule
    \midrule
    \multicolumn{2}{l}{Title: \href{https://www.gutenberg.org/ebooks/51597}{Gourmet} (\url{https://www.gutenberg.org/ebooks/51597})} \\
    \midrule
    \multicolumn{2}{l}{Q: What are some of the dishes that Bailey cooks for the crew?} \\ 
    \midrule
    The dishes Bailey cooks for the crew varies greatly, ranging from artificial vegetables to mock-meats. One dish that he makes is a mock-meat hamburger, with the pressed Chlorella tinted pink and seasoned by oregano and thyme. The dish is accompanied by dessert - a fudge made from dextrose-paste. More mock-meat dishes include a hamburger steak covered in a rich, meaty gravy lavishly seasoned with garlic. Another dish includes a mock individual head of lettuce dressed with vinegar and oil. The lettuce was made by Bailey constructing each synthetic lettuce leaf, with the narrator guessing the process to be out of pressing, rolling and shaping a green Chlorella paste. In contrast to some of the delicious dishes that Bailey makes, the Cook also delivers some less tasty meals in response to the Captain’s critiques. These included boiled Chlorella vulgaris in some soup and subpar algaeburgers. Bailey’s final dish in the story - and the best one yet - is an artificial steak that greets the crew with a barbecue smell. It is drenched with gravy and seasoned with a peppery and garlicy taste, and as the crew eats it, they find that the usually pond-scum taste that accompanies each repurposed chlorella meal is gone and instead, the taste and texture reflects actual steak. 
    & 
    One of the first-mentioned dishes that Bailey cooks is hamburger. He tries to create this out of the algae, seasoning the food to hide the flavors. He also serves a fudge for dessert that is compounded from the dextrose-paste of the carbohydrate recycler. After speaking with Paul initially, Bailey serves a dish of hamburger steak again. There is an individual head of lettuce served, along with a steak drenched in gravy. Later, he serves them a hot turkey supreme. The cheese-sauce is very believable, whereas the turkey is white and tender even though it is made from Chlorella. When Captain Winkelmann pushes Bailey too far, he begins to create disgusting foods. One of the first dishes he serves is boiled Chlorella vulgaris that resembles vomit. The coffee at noon also tastes of salt. However, at the very end of the story, Bailey succeeds in making his Chlorella steak actually taste like food.
    \\
    \midrule 
    Throughout their trip, Bailey does the best he can in order to replicate traditional food using the Algae. To impress the Captain, Bailey cooks a wide variety of foods including algae burgers, fudge, Steak with gravy and a head of lettuce, Hot turkey with cornbread and butter sauce, and medium rare steak. None of these foods impressed the Captain, so Bailey went back to cooking unappealing food such as a porridge-like broth and bad coffee. At the end, Bailey serves a new type of steak, which is hinted to be human steak from the Captain.
    & 
    Bailey made a lot of different dishes while working on the Sale ship. He cooked a hamburger and a fudge. He made a steak with rich meat gravy and lettuce, vinegar, and oil. An ersatz hot turkey supreme with a cheese sauce, cornbread, and a pottage was also served at some point. All of these were criticized by Captain Winkelmann. Mostly Bailey was working on the taste of steak, which at the end of the story, he managed to perfect to a certain extent, partly thanks to the captain’s constant remarks. 
    \\
    \bottomrule
    \end{tabularx}
    \caption{Additional example questions and reference summaries from \dataset.}
    \label{tab:dataset_examples_additional}
\end{table*}


\section{Human Evaluation}
\label{ax:human_eval}

We ask human raters to (re-)read the story, and then evaluate the quality of summaries along three axes:
\begin{itemize}
    \item Correctness: Presence of factual errors in responses, where a factual error is a statement that contradicts the story, or is not directly stated, heavily implied, or logically entailed by the story.
    \item Coverage: The degree to which the response contains all information and details from the story that are relevant to answering the question.
    \item Overall: Overall quality of the response, the primary considerations of which are the readability/intelligibility of the response, the correctness, and the coverage. We ask raters to use their best judgment in balancing these factors, as well as to incorporate other factors such as conciseness, repetitiveness, and copying.
\end{itemize}

We show the standard deviation of property ratings across questions in Table~\ref{tab:human_eval_std}.
\begin{figure*}[ht!]
     \centering
     \includegraphics[width=.99\textwidth]{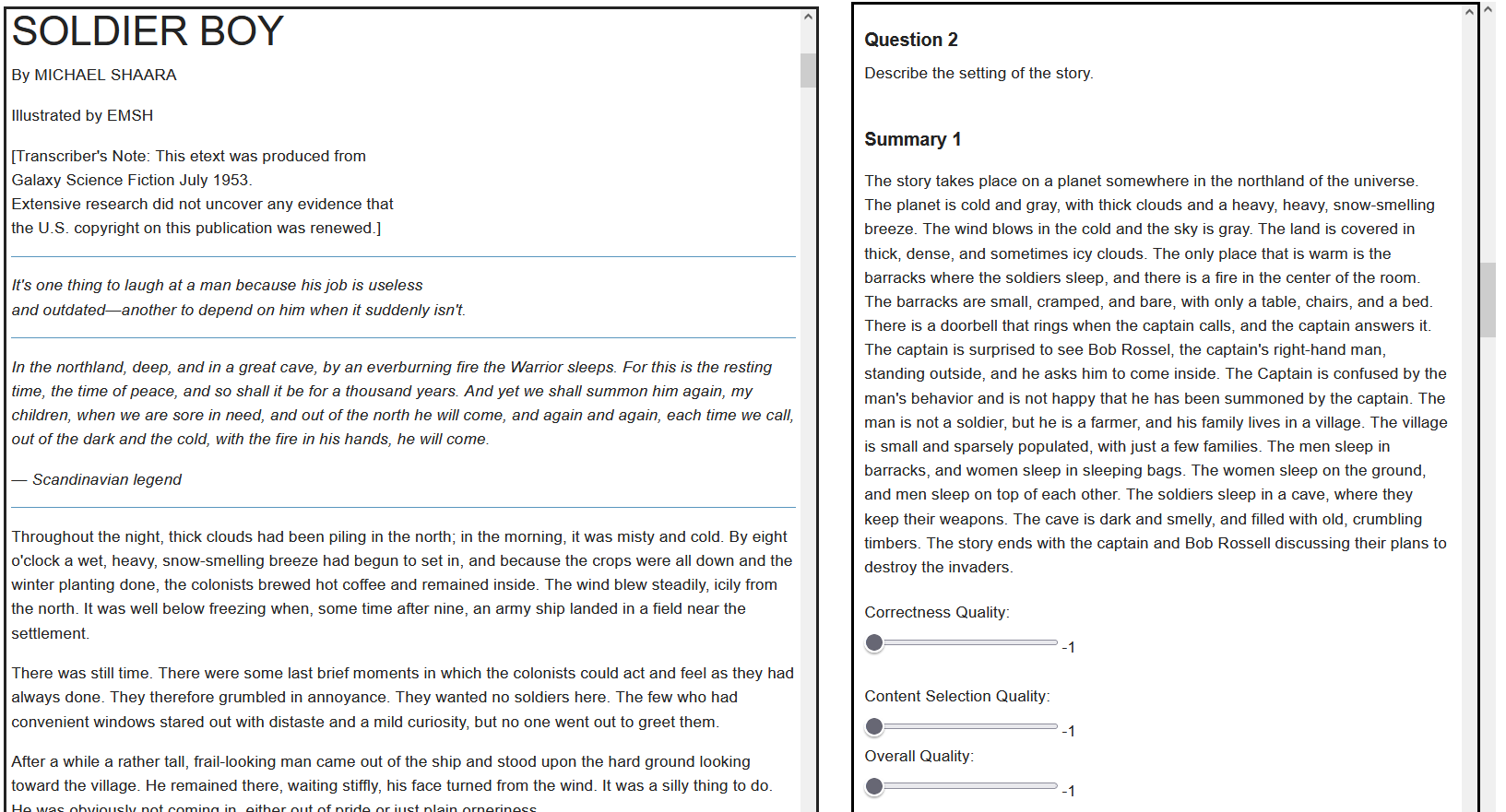}
    \caption{Screenshot of the human evaluation UI. Workers are shown the story on the left and five questions on the right. Each of the questions has three responses. For each response, the worker is instructed to rate the responses along the properties of correctness, coverage, and overall quality each along a scale of 1--100. Because the worker is shown three responses at a time, their ratings of each response induce a ranking over the responses. Additionally, workers are asked to highlight errors in responses in order to help them decide on the correctness property.}
    \label{fig:eval_ui}
    \vspace{-0.3cm}
\end{figure*}

\begin{table*}[t]
    \centering
    \begin{tabular}{lrrr}
    \toprule
    Model & Correctness & Coverage & Overall \\
    \midrule
    BART & 34.8$_{16.9}$ & 15.6$_{13.3}$ & 18.1$_{13.1}$ \\
    BART+DPR & 45.4$_{15.8}$ & 24.3$_{15.7}$ & 27.9$_{16.5}$ \\
    Human & 94.1$_{5.5}$ & 88.8$_{10.9}$ & 91.3$_{7.7}$ \\
    \bottomrule
    \end{tabular}
    \caption{Human evaluation results for two models and a human-written response. Ratings for each property are averaged across 3 workers, then averaged across questions. Standard deviation of property ratings across questions are shown in underscore.}
    \label{tab:human_eval_std}
\end{table*}

\end{document}